\documentclass[letterpaper, 10 pt, conference]{ieeeconf}  

\IEEEoverridecommandlockouts                              

\usepackage{graphicx}
\usepackage{subfigure}
\usepackage{amsmath}
\usepackage{amssymb}
\usepackage{bm}
\usepackage{xcolor}
\usepackage{physics}
\usepackage{hyperref}
\usepackage{microtype}
\usepackage[some]{background}

\usepackage[acronym]{glossaries}

\let\mathbf\symbf

\newacronym{EV}{ev}{EGO vehicle}

\renewcommand{\baselinestretch}{0.98}

\BgThispage
\backgroundsetup{
    scale=0.9,
    color=black,
    opacity=0.6,
    angle=0,
    position=current page.north,
    vshift=-1.5cm,
    hshift=0cm,
    contents={%
        \begin{minipage}{\paperwidth}
            Published at 2024 IEEE International Conference on Robotics and Automation (ICRA 2024):\\
            © 2024 IEEE.  Personal use of this material is permitted.  Permission from IEEE must be obtained for all other uses, in any current or future media, including reprinting/republishing this material for advertising or promotional purposes, creating new collective works, for resale or redistribution to servers or lists, or reuse of any copyrighted component of this work in other works.
        \end{minipage}
    }
}

\title{\LARGE \bf

Are you a robot? Detecting Autonomous Vehicles from Behavior Analysis
}

\author{Fabio Maresca$^{*}$, Filippo Grazioli$^{*}$, Antonio Albanese$^{\Psi}$, \\Vincenzo Sciancalepore$^{*}$, Gianpiero Negri$^{\dag}$, Xavier Costa-Perez$^{\ddagger}$
\thanks{$^{*}$F. Maresca, F. Grazioli and V. Siancalepore are with NEC Laboratories Europe GmbH
        {\tt\small \{name.surname\}@neclab.eu}}%
\thanks{$^{\Psi}$A. Albanese is with Flyhound Co.
        {\tt\small antonio.albanese@flyhound.com}}%
\thanks{$^{\ddagger}$X. Costa-Perez is with i2CAT Foundation, NEC Laboratories Europe GmbH and ICREA.
        {\tt\small xavier.costa@icrea.cat}}%
\thanks{$^{\dag}$G. Negri is with Amazon Global Robotics - EU Innovation Lab.
        {\tt\small ngianpie@amazon.com}}
\thanks{This work was partially supported by EU FP for Research and Innovation Horizon 2020 under Grant Agreements No. 956670 (5GSmartFact) and No. 101139130 (6G-DISAC).}%
}

\begin{document}


\maketitle
\thispagestyle{empty}
\pagestyle{empty}

\begin{abstract}

The tremendous hype around autonomous driving is eagerly calling for emerging and novel technologies to support advanced mobility use cases. As car manufactures keep developing SAE level 3+ systems to improve \emph{the safety and comfort} of passengers, traffic authorities need to establish new procedures to manage the transition from human-driven to fully-autonomous vehicles while providing a feedback-loop mechanism to fine-tune envisioned autonomous systems. 
Thus, a way to automatically profile autonomous vehicles and differentiate those from human-driven ones is a \emph{must}. 

In this paper, we present a fully-fledged framework that monitors active vehicles using camera images and state information in order to determine whether vehicles are autonomous, without requiring any active notification from the vehicles themselves. Essentially, it builds on the cooperation among vehicles, which share their data acquired on the road feeding a machine learning model to identify autonomous cars. We extensively tested our solution and created the NexusStreet dataset, by means of the CARLA simulator, employing an autonomous driving control agent and a steering wheel maneuvered by licensed drivers. Experiments show it is possible to discriminate the two behaviors by analyzing video clips with an accuracy of $\sim\! 80\%$, which improves up to $\sim\! 93\%$ when the target's state information is available. Lastly, we deliberately degraded the state to observe how the framework performs under non-ideal data collection conditions.
\end{abstract}

\glsresetall
\section{Introduction}
With the advent of emerging communication technologies, the automotive sector has exhibited relevant industrial profits towards the so-called revolution of \emph{connected and autonomous cars}. This digital transformation enables advanced services drawing attention to two crucial societal aspects, such as safety of road users and sustainability of involved infrastructure, thereby delaying the transition to connected roads with fully-autonomous cars as part of the SAE level 3+: this implies vehicles implementing an environmental perception module~\cite{duarte2018impact}.
Within this transition phase, it appears clear a need for the coexistence of human drivers and autonomously-driven vehicles: computers will take over the maneuvering control of the cars to take specific actions (e.g., highway driving or parking), 
while giving it back upon critical situations~\cite{virdi2019safety}.  

\begin{figure}[t]
    \centering  
    \includegraphics[clip,width =  \linewidth]{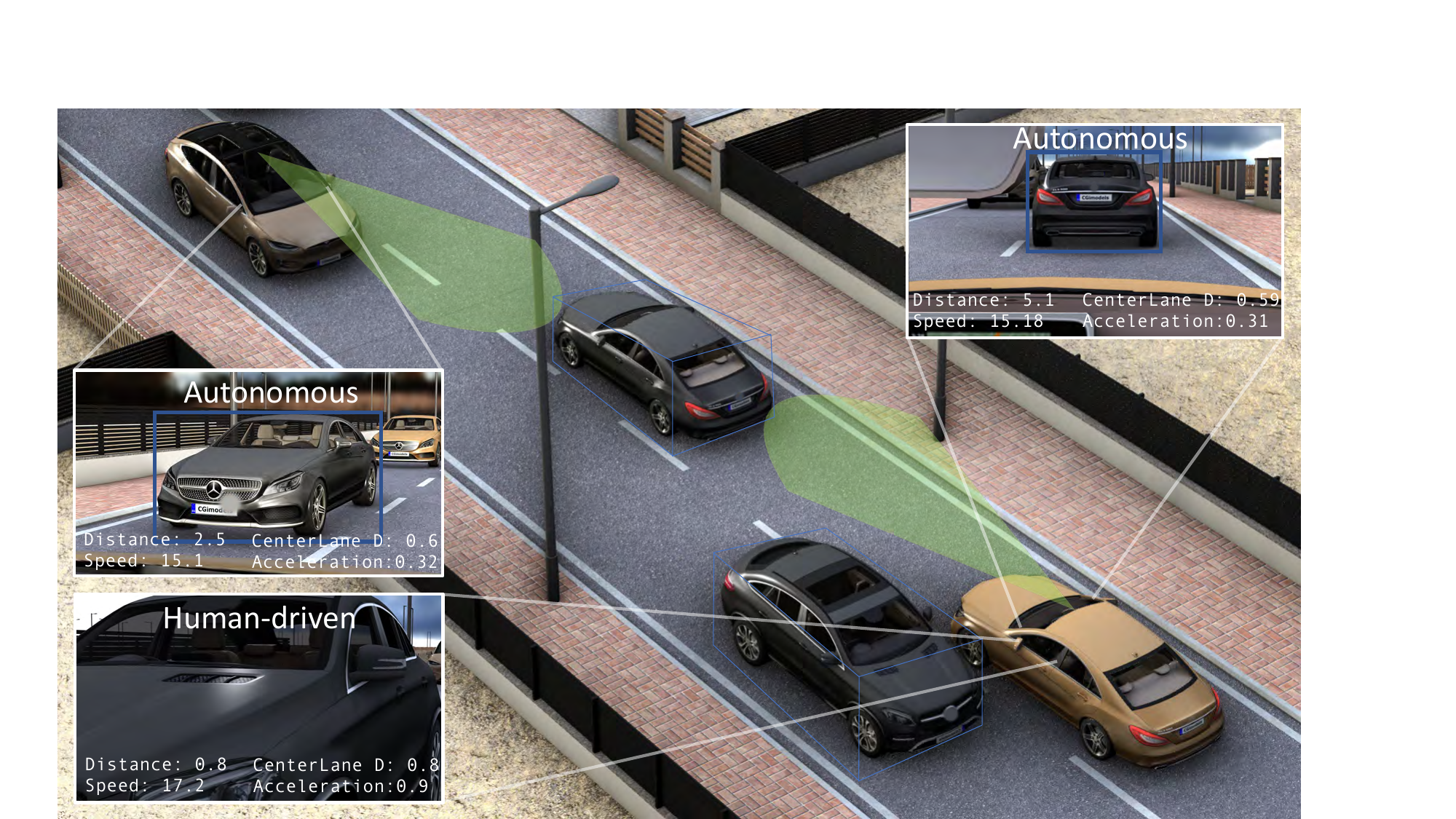}
    \caption{Envisioned vehicular communication scenario. Vehicles report state information and collectively identify autonomous cars.}
    \label{fig:scenario}
\end{figure}



Such mixed traffic conditions~\cite{parth2017} will involve a number of severe safety issues that must be tackled~\cite{9636795}: autonomously-driven vehicles will slowly build the \emph{risk perception}~\cite{peng2023sotif} exacerbating safeguarding concerns; it is crucial to spot autonomous cars with (potential) inconsistent behaviors~\cite{hussain2022deepguard}. We propose an overarching framework to automatically detect whether a vehicle is autonomous or human-driven based on gathered behavioral features and to provide feedbacks for improving autonomous systems: it benchmarks established machine learning models trained on temporal series of images and additional numeric features as the state information of the target vehicle to classify.\\
To tackle this problem, we built and exploited a dataset that collects human and autonomous driving scenes, in the form of videos, i.e., sequences of RGB images. These are recorded by the EGO vehicle (EV) following a {\it target vehicle} (TV), whose state, collected alongside, is characterized by speed, acceleration, rotation, distance from the EV and distance from lane center. We have built the dataset using the CARLA simulator~\cite{Dosovitskiy17} and the Baidu Apollo autonomous driving agent to control the target in the autonomous settings~\cite{huang2018apolloscape, ap}, whereas the human driving behavior has been acquired by manually steering the TV with a Logitech G29 steering wheel. The dataset is dubbed as NexusStreets\footnote{{\it NexusStreets} is inspired by the Nexus series of replicants in Blade Runner, who were designed to be nearly indistinguishable from humans. The name emphasizes the idea of a mix of human-driven and autonomous vehicles on the same streets. 
} and is publicly released on Zenodo at~\href{https://zenodo.org/records/7682484}{https://zenodo.org/records/7682484}.



\subsection{Related Work}

A useful taxonomic characterization of behavior prediction methods is based on the type of input features employed by the models \cite{mozaffari2020deep}.
%
In~\cite{zyner2019naturalistic, xin2018intention}, the authors propose machine learning models that predict vehicle behavior from the track history of the TVs states. The accuracy of such methods is limited by the performance of the EV's perception module. Note that they do not consider vehicle-vehicle, nor vehicle-environment interaction patterns. In~\cite{piroad_infocom21} the authors propose a deep learning framework to automatically learn regular mobile traffic patterns along roads thereby detecting non-recurring events and classify them by severity level to assist the emergency operations and support advanced services provided by the network operator. 


When other surrounding vehicles' track histories are considered, patterns are extracted from the interactions between them and the target vehicles~\cite{diehl2019graph,  ma2019trafficpredict}. Analogously to the works presented above, these methods' performance is limited by the accuracy of the EV's perception that provides the track histories. However, they can learn from a richer input representations and hence capture more complex dynamics.

\begin{figure}[t]
    \centering  
    \includegraphics[clip,width =  \linewidth]{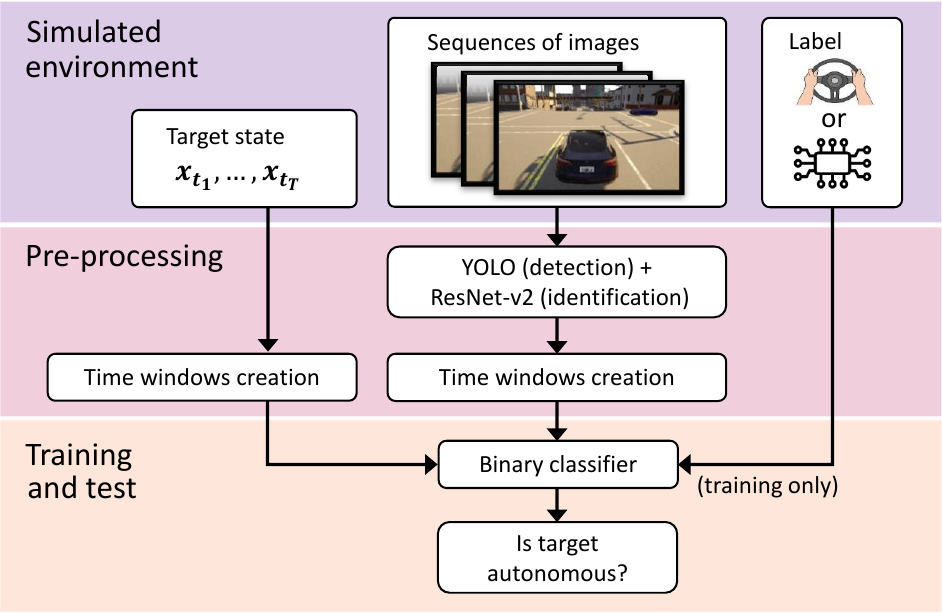}
    \caption{Illustration of the Autonomous Vehicle Detection Methodology}
    \label{fig:full-pipeline}
\end{figure}

To account for vehicle-environment interactions, a further class of methods inputs simplified bird's eyes views of the scene~\cite{lee2017convolution, schreiber2019long}.
%
Finally, \cite{luo2018fast, casas2018intentnet} propose to directly predict vehicle behavior from the raw sensor data to overcome the information loss injected by the perception stack.

To the best of our knowledge, no existing work investigates the problem of spotting autonomous vehicles from observations providing feedbacks on inconsistent behaviors.

\subsection{Contributions}
The paper contributions can be summarized as follows: $i$) we generate a novel dataset combining autonomous and human driving scenes, $ii$) we present benchmark results of various machine learning models on the driving scene classification task, $iii$) we model the problem as a supervised classification of multivariate time series describing the temporal evolution of the Target Vehicle (TV) state, including a second multivariate time series representing 2D bounding box detection of the TV in the pixel coordinates space over time, $iv$) we carry out an exhaustive simulation campaign to validate our framework, $v$) we make the NexusStreets dataset publicly available to the scientific community. Fig.~\ref{fig:full-pipeline} provides a schematic illustration of the full pipeline.

Our experiments show that it is possible to discriminate with high accuracy ($>0.95$ auROC) whether a TV is autonomous or human-driven alone from the observation of its driving behavior in the form of state and 2D object detection.

\section{Dataset Generation}
\subsection{Simulation Environment and Considered Scenarios}
Autonomous and manual driving scenes have been collected using a nearly-realistic simulator, namely CARLA~\cite{Dosovitskiy17} used in driving-related research. 
It offers access to the vehicles' sensors and formally describes the roads according to OpenDRIVE specifications~\cite{dupuis2010opendrive}. 
It also provides interfaces with external tools and software, e.g., ROS~\cite{quigley2009ros, macenski2022robot}, and autonomous driving agents to control vehicles. Additionally, it allows the integration of external steering wheels to directly drive the vehicles: human behaviors have been reproduced with a Logitech G29 steering wheel, which supports realistic force feedback and first-person driving experience.

Five default maps are available offering a variety of streets, from narrow to larger ones, up to a $5$-lanes highway. Each map is characterized by different curves, traffic signs and lights dispositions. 
For each city, we considered different levels of traffic, spanning from no traffic to heavy traffic conditions, where a hundred non-player characters (NPCs) spawn randomly in the simulated environment. 
Every scene is repeated with $10$ different weather conditions 
allowing to acquire video sequences in a richer spectrum of light and visibility conditions\footnote{The weather conditions do not affect the physics of the simulator, nor the actors and NPCs' behaviors. Different weather conditions have been considered to make the detection task (section~\ref{section-detection}) closer to the reality.}. 


\begin{figure}[t]
    \centering  
    \includegraphics[clip,width =  \linewidth]{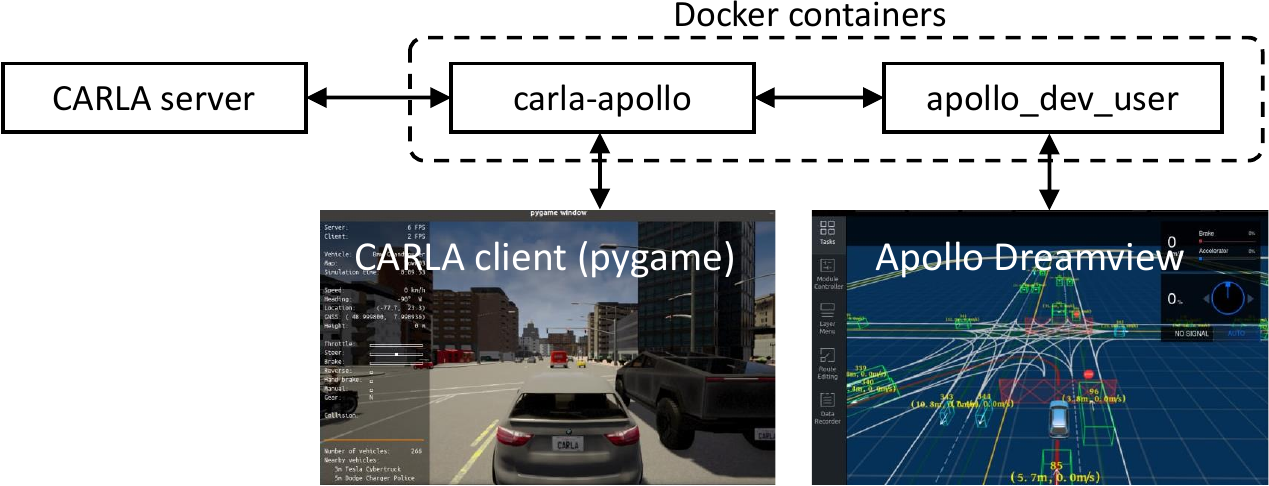}
    \caption{Schematic illustration of the {\tt carla-apollo} bridge architecture}
    \label{fig:carla-apollo_bridge_architecture}
\end{figure}



\subsection{Description of the Main Actors}
We assume in our dataset two main actors:
\begin{itemize}
    \item {\bf Target vehicle (TV)}, the leading vehicle for which we acquire state information and whose behavior we aim to classify;  
    \item {\bf EGO vehicle (EV)}, the vehicle following and monitoring the TV. It records the scenarios with a monocular RBG front-camera and, ideally, it retrieves and computes data by means of additional sensors (radar, LiDAR, etc.) and/or external sources, e.g., other communicating vehicles and objects in smart cities.
\end{itemize}

\begin{figure}[t!]
    \centering  
    \includegraphics[clip,width =  \linewidth]{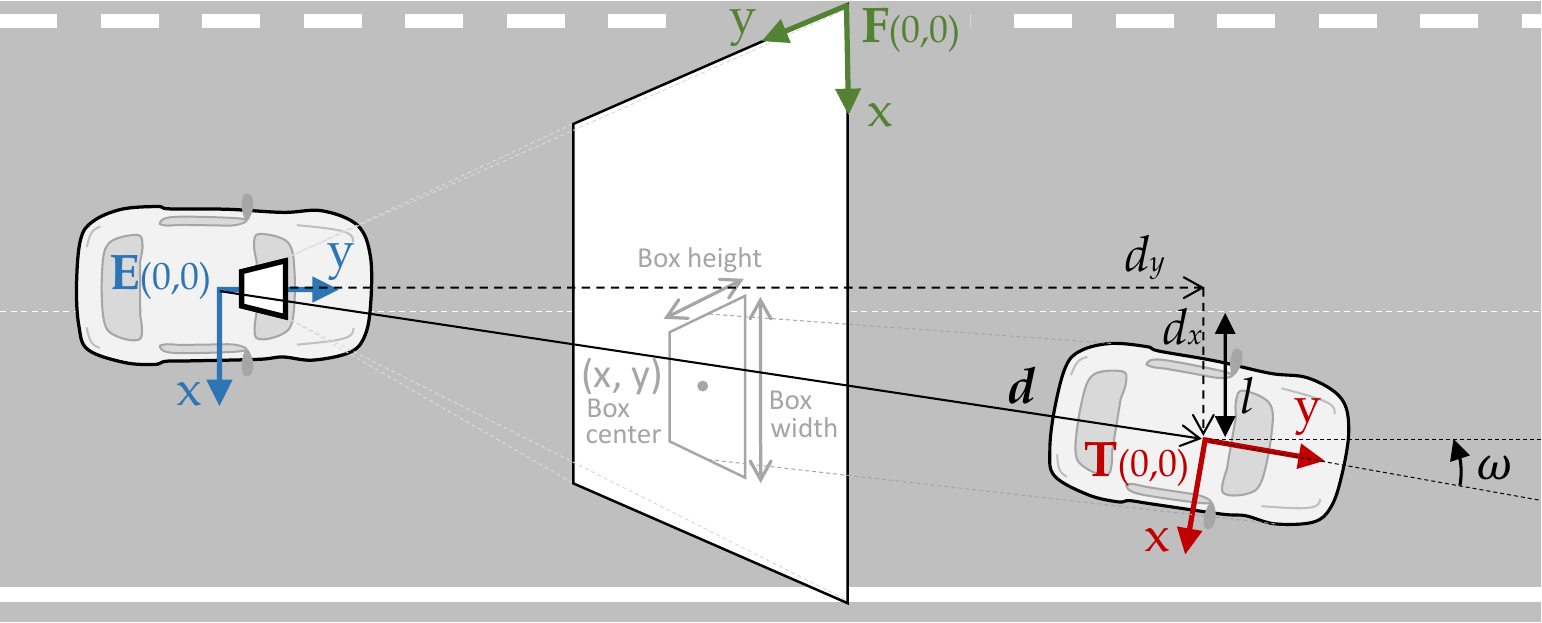}
    \caption{$T$ and $E$ are the two reference systems for TV and EV, respectively. Their origins lay in their center of mass. For simplicity, we only depict the $x$- and $y$-axis, however the $z$-axis is also considered in the simulations.
    The TV state is constituted by its position vector $\bm{d}$, speed $\dot{\bm{d}}$ and acceleration $\ddot{\bm{d}}$ w.r.t. the $E$ frame, as well as the distance from the lane center $l$ and its yaw angle $\omega$ calculated in $T$.
    Moreover, we use the reference system $F$ to represent the 2D bounding box coordinates. $F(0,0)$ corresponds to the upper left pixel of each 960x540 JPEG image acquired by the EV camera.
    }
    \label{fig:ego-test-vehicles}
    \end{figure}

\begin{figure*}[t]
\centering
\includegraphics[width=\textwidth]{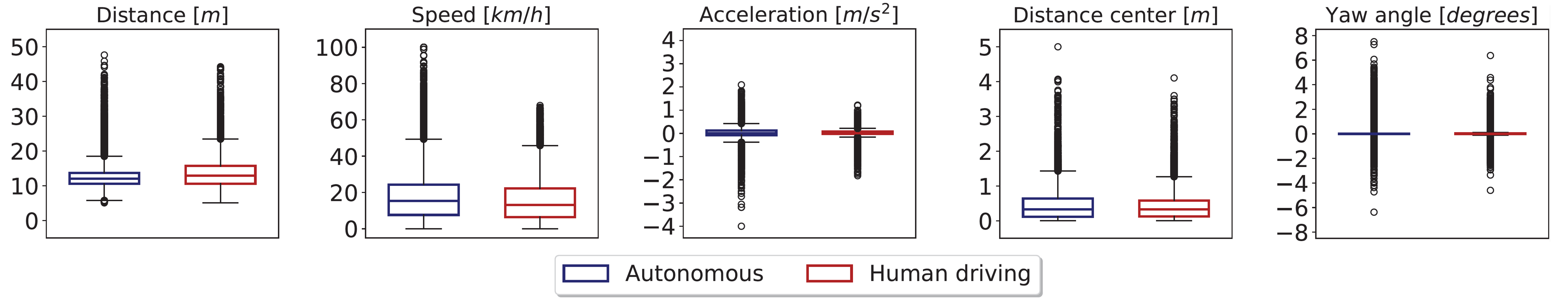}
\caption{Distribution of each TV state feature, highlighting $25$th quartile, median, $75$th quartile, and values falling outside such two boundaries.}
\label{fig:features_boxplot}
\end{figure*}

\noindent The dataset properly combines autonomous driving scenes wherein the TV is implemented using the Baidu Apollo~\cite{huang2018apolloscape, ap} autonomous driving agent, and human-driven scenes wherein the TV is controlled by skilled people directly driving with a steering wheel in a first-person perspective~\cite{haris2022navigating}. 


The EV follows the TV via a PID controller that mimics the TV's path by automatically adjusting its speed basing on the distance from the TV. 

\subsection{Sampling Scenes and Target States}
The dataset consists of 520 scenes, i.e., 260 pairs of scenes where the TV is controlled by the autonomous agent and the human driver in  identical scenarios.
Two $60$s runs for every (city, traffic, weather) combination result in nearly $9$ hours of total driving time.
For each timestamp, the $960\times 540$ RGB image captured by the EV's front camera is acquired, together with the TV state provided by the simulator.

Each video is truncated at $60$s to ensure uniform duration. Additionally, we down-sample the scenes and the acquired states to the minimum sampling rate, guaranteeing a correspondence between the images and the state at all times. This results in $\sim\!2$fps and all scenes present $120$ timestamps.

At each sampling timestamp, the EV captures the TV state $\bm{x^S} = \big[\|\bm{d}\|_2, \|\dot{\bm{d}}\|_2, \|\ddot{\bm{d}}\|_2, l, \omega \big]^T$, where $\bm{d} \in \mathbb{R}^3$ is the position vector of TV in EV-centered reference system $E$, $\|\cdot \|_2$ denotes the $L^2$-norm of the vector, $l\in\mathbb{R}$ is the distance of the TV from the lane center w.r.t. the TV-centered reference system $T$ and $\omega\in \mathbb{R}$ is the TV's yaw angle. We denote the time derivative with the dot-notation, i.e., $\dot{\bm{d}}$ is the speed vector and $\ddot{\bm{d}}$ is the acceleration. Fig. \ref{fig:ego-test-vehicles} provides a depiction of the $E$ and $T$ reference systems.

\begin{figure*}
\centering
\subfigure{
\centering
\includegraphics[height=3.3cm]{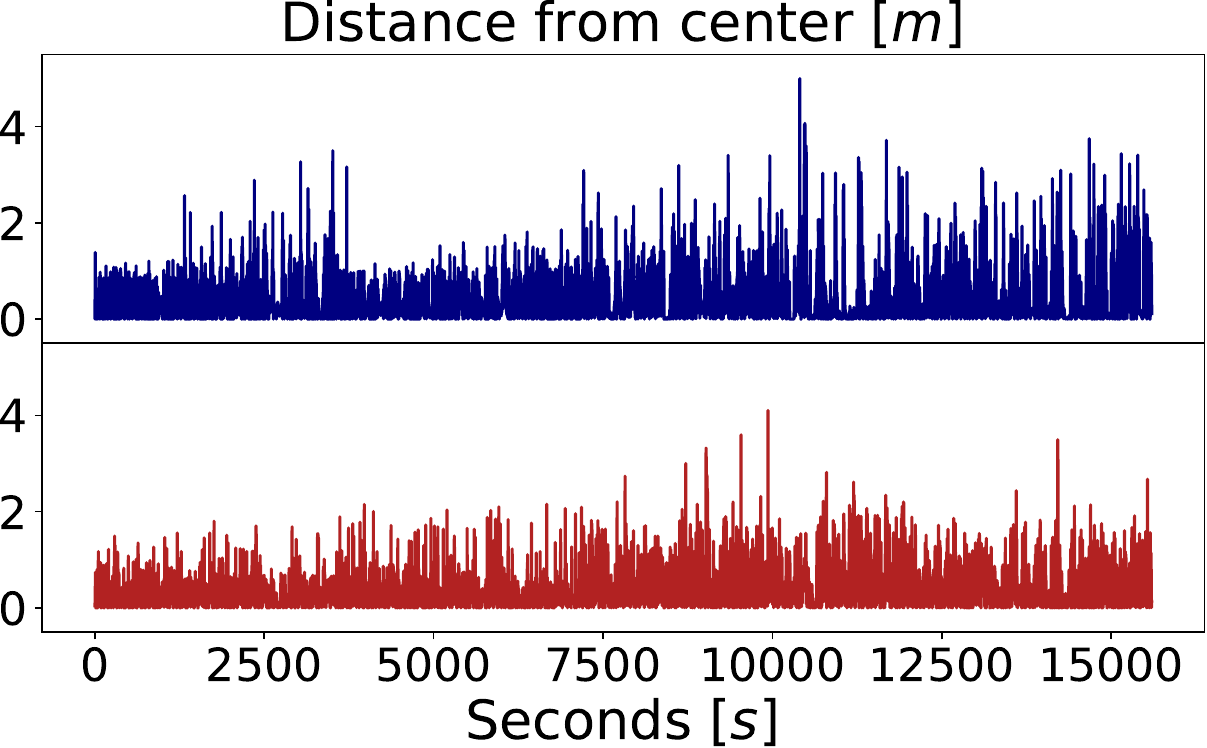}
}
\subfigure{
\centering
\includegraphics[height=3.3cm]{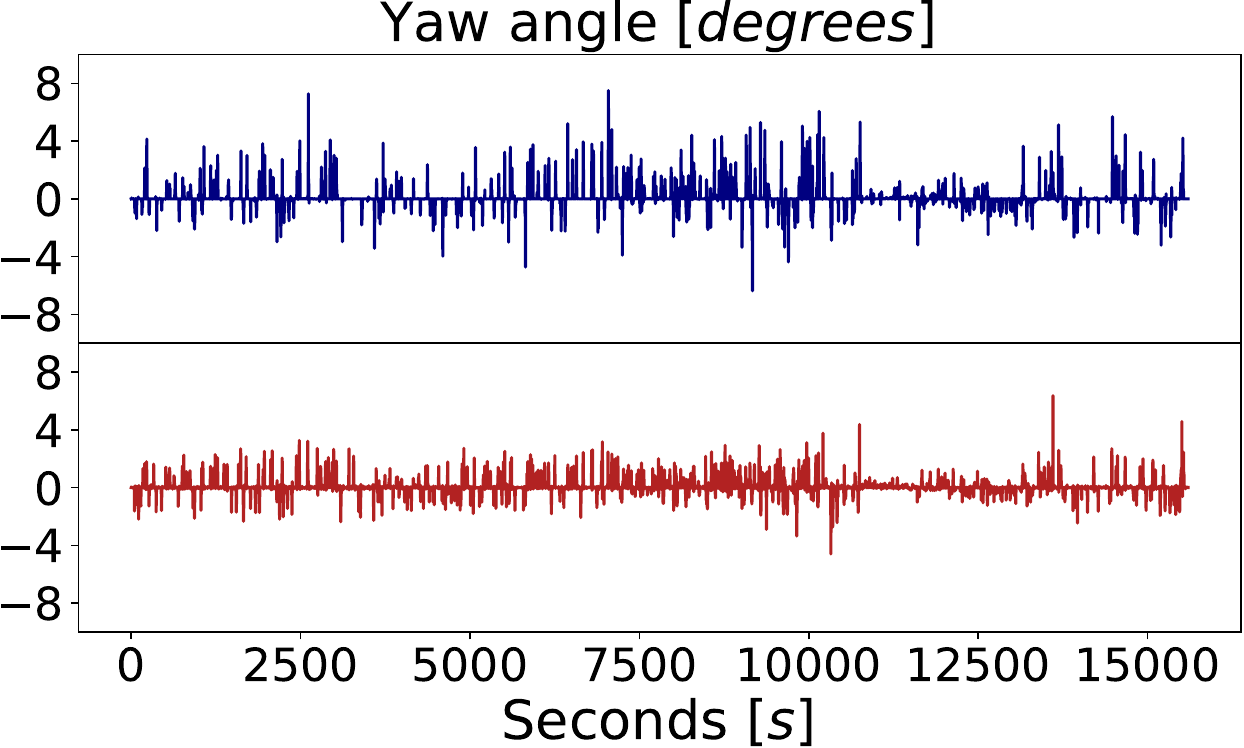}
}
\subfigure{
\centering
\raisebox{2\height}{
\includegraphics[width=.16\textwidth]{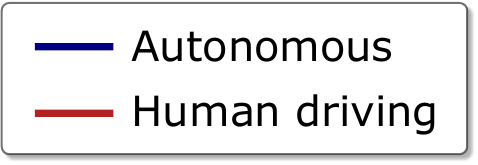}
}}
\caption{Trends of the state features over time. The maximum value $15,600$s on the x-axis corresponds to the concatenation of $260$ scenes lasting $60$s each.}
\label{fig:features}
\end{figure*}

While the TV state $\bm{x^S}$ is provided by the simulator, in real-life scenarios it could be estimated from the EV's perception stack, or also shared by other surrounding vehicles and objects by means of Vehicle-to-Everything (V2X) communications. Interestingly, multiple followers can monitor and profile the same target vehicle while exchanging acquired information. 
Figs.~\ref{fig:features_boxplot} and \ref{fig:features} depict the distributions of the features that constitute the TV state.

\begin{figure*}
\centering
\subfigure{
\includegraphics[height=3.3cm]{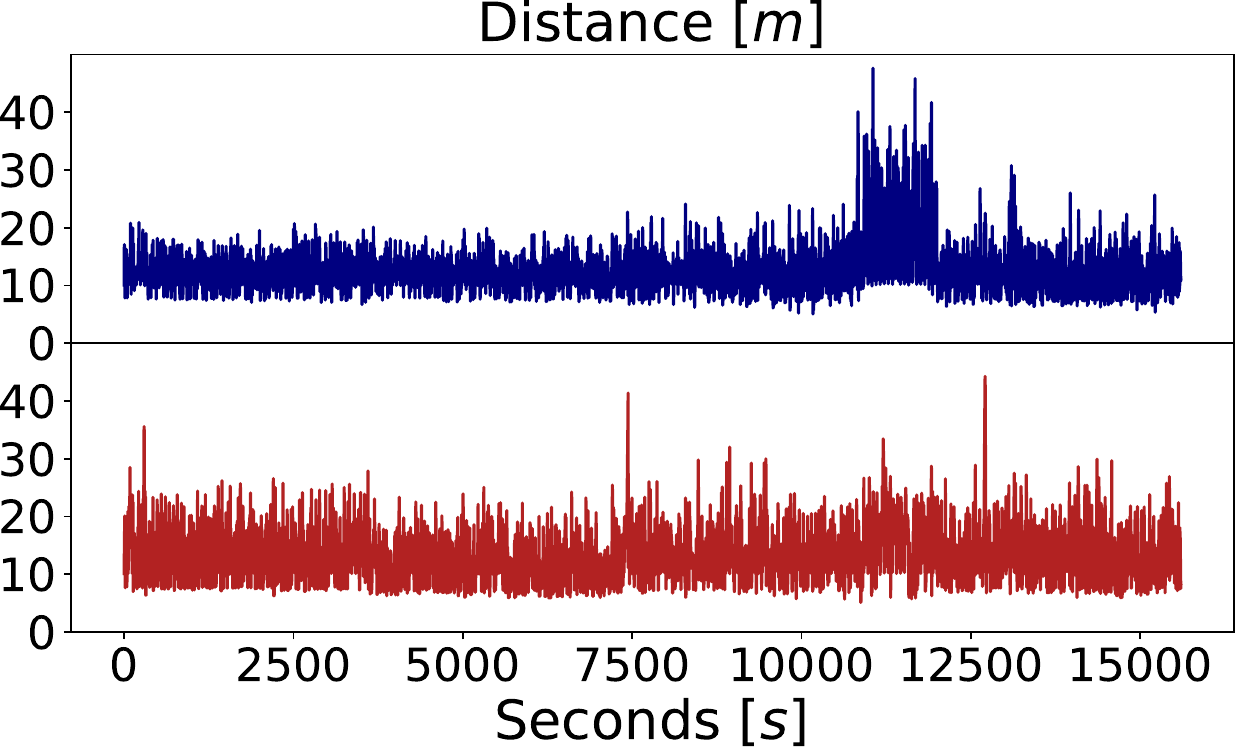}
}
\subfigure{
\includegraphics[height=3.3cm]{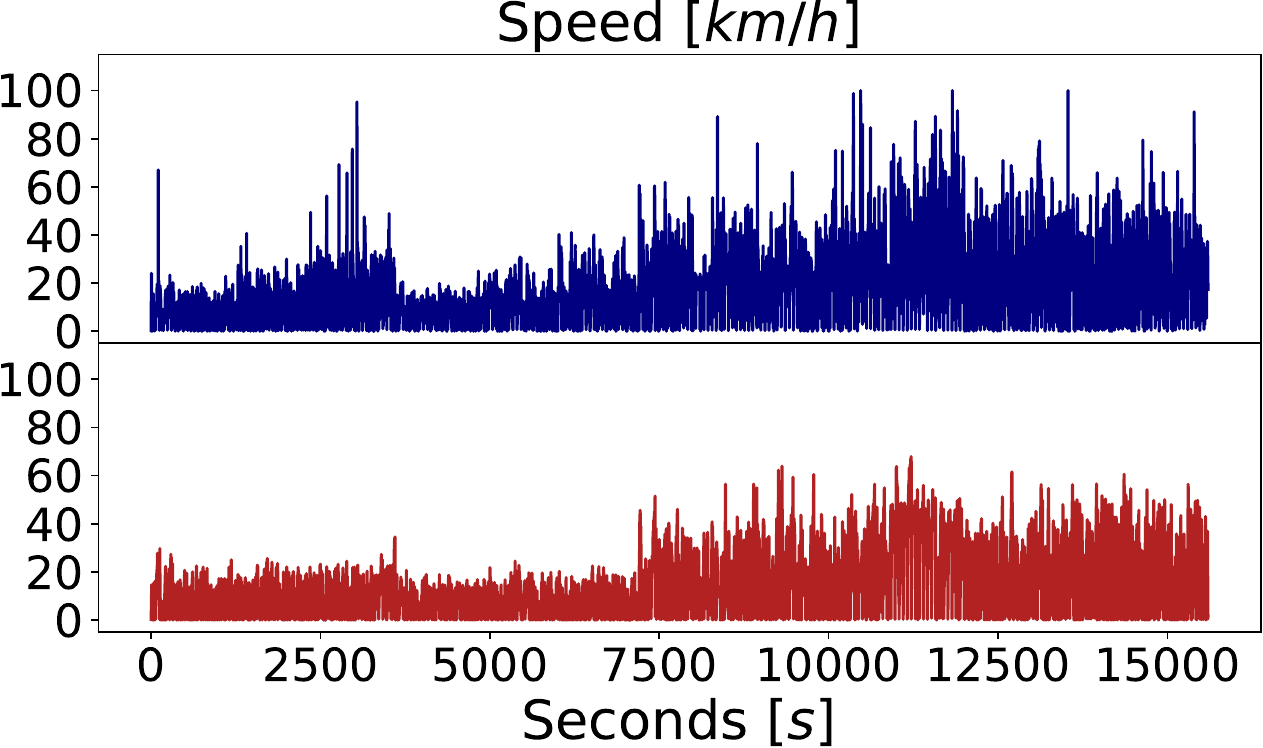}
}
\subfigure{
\includegraphics[height=3.3cm]{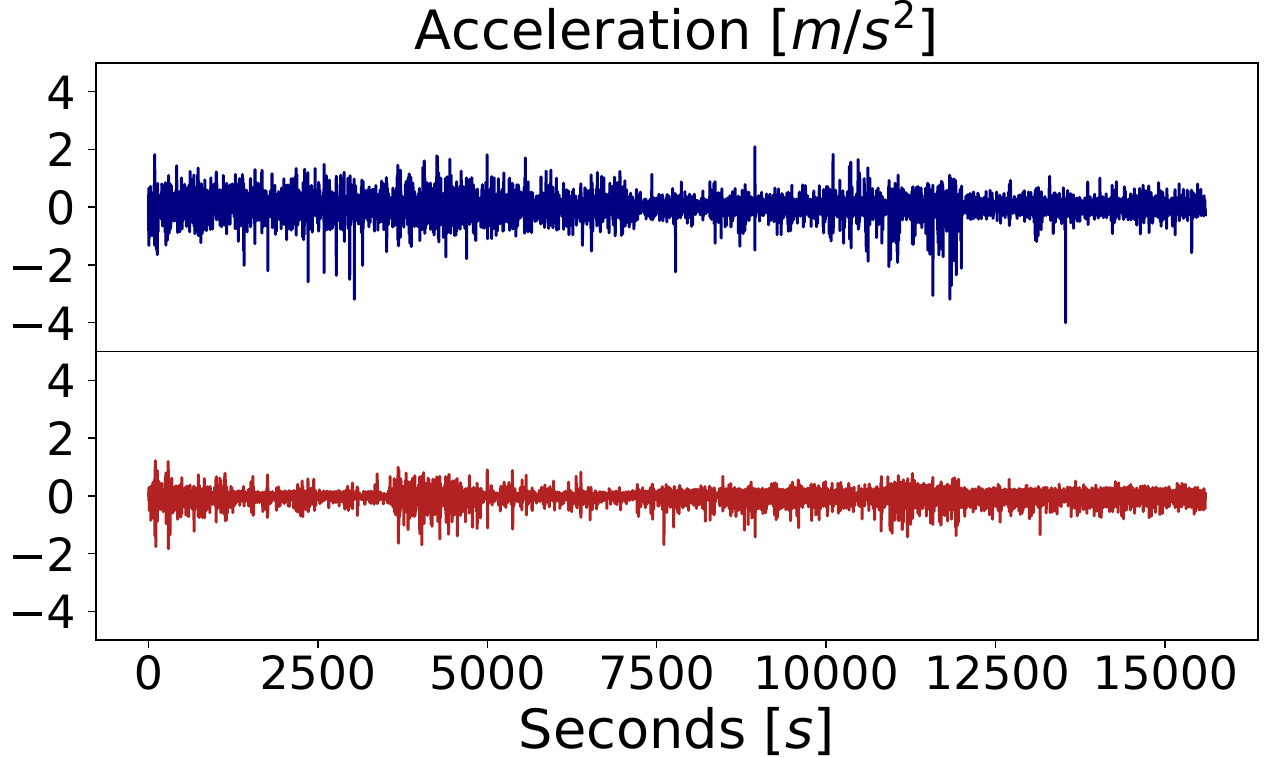}
}
\end{figure*}

\subsection{2D Target Vehicle Detection}
\label{section-detection}
In addition to the TV state, we provide 2D object detections in the form of bounding box coordinates w.r.t the reference system $F$ at each timestamp (see Fig.~\ref{fig:ego-test-vehicles}).
We operate 2D object detection using a pre-trained YOLOv7 \cite{wang2023yolov7} model.
This convolutional neural network (CNN) outputs bounding boxes and class probabilities for each object detected in the image. 
For each acquired frame, we consider objects labeled as ``cars'' with a confidence $\geq 0.9$ and ignore other classes. We store the detection as vectors $\bm{x^D} = \big[x, y, h, w \big]$, where $(x,y)$ are the bounding box center coordinates, $h$ its height and $w$ its width.
For each frame, multiple vehicles are detected. It is therefore necessary to identify the bounding box of the TV.
To achieve this goal, for each weather and light condition, we first collect a set of $2$ images (as template) depicting the TV rear and side.
Then, we embed the template images using a ResNet-50v2~\cite{he2016deep} model trained on ImageNet~\cite{deng2009imagenet}. Embedding vectors are obtained from the activation of the final pre-softmax fully-connected layer.
Afterwards, for each frame, we embed all detection crops and then compute their average Euclidean distance from the TV template.
The detection crop whose embedding presents the smallest average Euclidean distance is designated as the true TV detection\footnote{Analogously, to a real-world perception module, this approach is not exempted from noise. Nevertheless, the combination of YOLO and ResNet-50v2-based template matching is able to detect the TV most of the times. Mistakes are considered outliers given their small numbers.}.



\section{Experiments and Discussion}
Hereafter, we show experimental results for the vehicle behavior classification task, which is modeled as a supervised learning problem. We then present empirical results for the future state prediction process, which is modeled as an autoregressive problem.

\subsection{Vehicle Behavior Classification}
We first present a formalization of the problem, then we compare different machine learning models performances for the behavior classification task. Specifically, this is based on the history of the target vehicle (TV) state, 2D detection in the pixel coordinates space and a combination of both.
Finally, we show how information loss in the TV state leads to a performance degradation at the classifier. 

{\bf Problem formulation.} 
Formally, for a given scene $i$, let us define the whole history of TV state vectors as the following:
\begin{equation}\label{eq:samples-s}
    \bm{X}^S_i = \Big[\bm{x}^S_{i,t_1}, \dots, \bm{x}^S_{i,t_{T_i}} \Big]^T \in \mathbb{R}^{T_i \times 5} \:\:,
\end{equation}
where $\bm{x}^S_{i,t}$ is the state vector at timestamp $t$ and $T_i$ represent the number of available timestamps. Analogously, let us define the whole history of 2D detections for a given scene as the following
\begin{equation}\label{eq:samples-d}
    \bm{X}^D_i = \Big[\bm{x}^D_{i,t_1}, \dots, \bm{x}^D_{i,t_{T_i}} \Big]^T \in \mathbb{R}^{T_i \times 4} \:\:.
\end{equation}

To each scene, we assign a binary label $y_i \in \{0, 1\}$, whose value represents whether the TV is classified as autonomous or human-driven.

Given a total number of scenes $N$, we can now define two datasets, i.e., the state information and the 2D detection information in the pixel coordinates space, accordingly: 
\begin{align}\label{eq:datasets}
\mathcal{X}^S = \Big\{\Big(\bm{X}^S_i, y_i\Big)\Big\}^N_{i=1}; \qquad
\mathcal{X}^D = \Big\{\Big(\bm{X}^D_i, y_i\Big)\Big\}^N_{i=1}.
\end{align}
Analogously, considering the concatenation of the state vectors with the 2D detections, we can define a joint dataset as follows:
\begin{equation}\label{eq:joint-dataset}
\mathcal{X}^{S+D} = \Big\{\Big(\Big[\bm{X}^S_i, \bm{X}^D_i\Big], y_i\Big)\Big\}^N_{i=1}.
\end{equation}

For the three settings considered, i.e., state-only (S), detection-only (D) and both (S+D), our goal is to estimate the true probability distribution $p(y | \bm{X})$ by a parametric approximation $q_{\bm{\theta}}(y | \bm{X})$, such that
\begin{align}\label{eq:prob_dist}
	q_{\bm{\theta}}(y | \bm{X}) \approx p(y | \bm{X}), 
\end{align}
where $\bm{\theta}$ is the set of parameters. Hereafter, we describe the two machine learning models employed in our experiments for the estimation of $q_{\bm{\theta}}(y | \bm{X})$.

{\bf Machine Learning Models.} 
We benchmark two different machine learning models, namely a Random Forest (RF)~\cite{ho1995random} and a deep neural network based on Long Short-Term Memory (LSTM) cells~\cite{hochreiter1997long}.

\begin{figure}[t]
    \centering  
    \includegraphics[clip,width =  \linewidth]{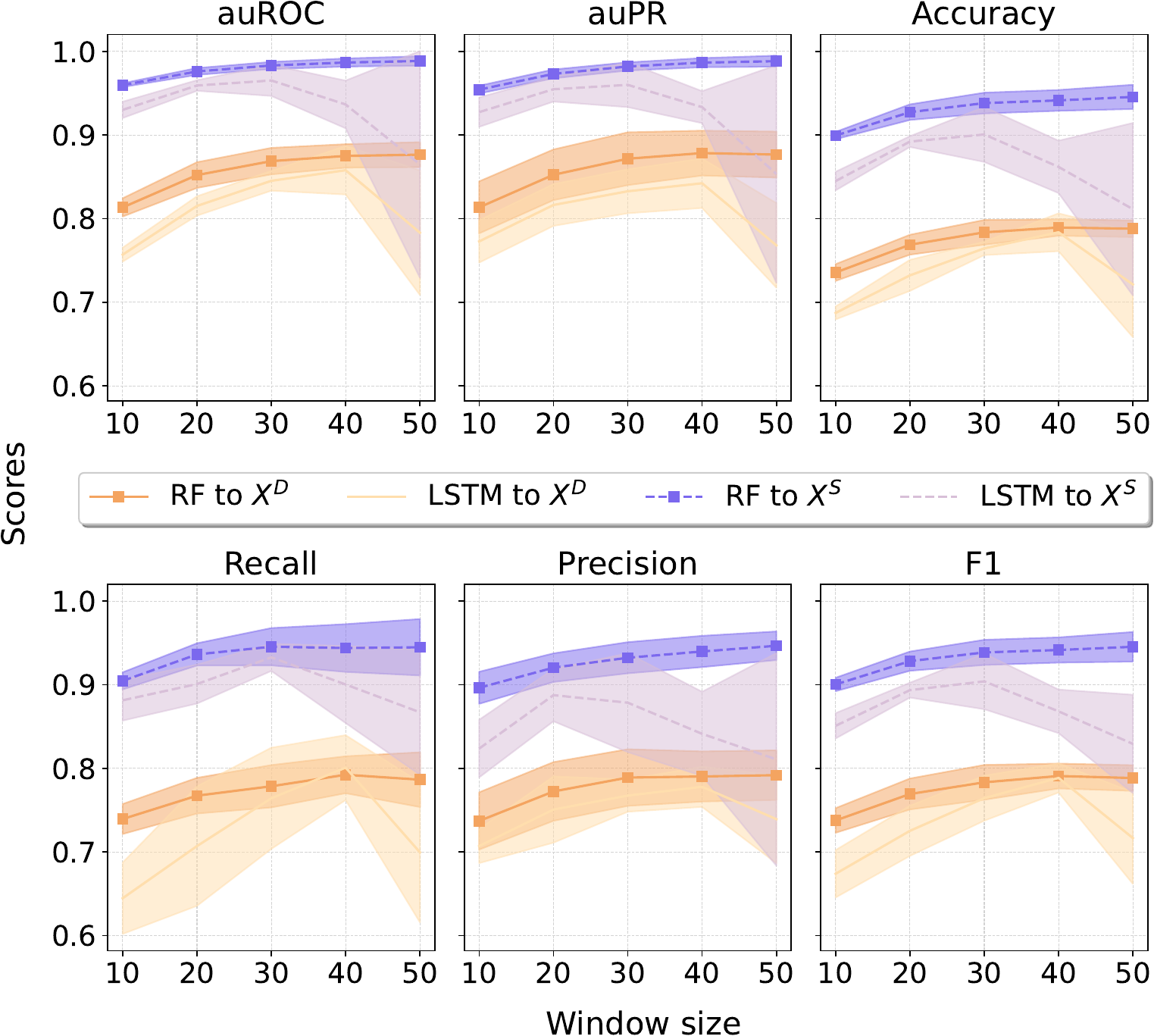}
    \caption{Vehicle behavior classification results. These have been obtained by applying RF and LSTM models to the datasets $\mathcal{X}^S$ and $\mathcal{X}^D$ separately. Reported values are test mean scores over $5$ repeated experiments. Confidence intervals are standard deviations.}
    \label{fig:results_RF-LSTM}
\end{figure}

For both models, we perform hyper-parameter optimization via a $5$-fold stratified cross-validated grid search.
Given a training/test split, only the training set is used for the hyper-parameter optimization in order to avoid information leak.
Each hyper-parameter configuration obtains $5$ validation scores. The configuration that achieves the maximum average area under the receiver operator characteristic curve (auROC) is selected. A new model with the optimal configuration is trained on the training set and tested on the left-out test set.

For the RF, we optimize the number of trees over $\{100k \:|\: 1 \leq k \leq 10\}$, the minimum number of samples required to be at a leaf node over $\{k \:|\: 1 \leq k \leq 5\}$, the function to measure the quality of a split over $\{\text{Gini impurity}, \text{Shannon entropy}\}$ and the number of features to consider when looking for the best split over $\{\sqrt{n}, \log_2 n\}$, where $n$ is the number of input features. Since the RF inputs vectors, we unroll the input matrices $\bm{X}$ to $1$D vectors.

For the LSTM-based model, we optimize the number of stacked LSTM cells over $\{ 1, 4, 8\}$, the drop-out rate over $\{0.1, 0.3\}$ and the output dimension of the LSTM cells over $\{32, 64, 128\}$.
The LSTM-based network is trained for a maximum of $1000$ epochs by optimizing a binary cross-entropy loss, employing the Adam optimizer and early stopping with $5$ epochs patience to monitor the validation auROC. The batch size was set to $256$ and the learning rate to $10^{-4}$. ReLU activations have been employed in the model and a final sigmoid function was used for the output.

{\bf Performance Evaluation.} 
With the goal of providing an unbiased estimate of the models classification performance, we perform $5$ independent random training/validation/test splits.
For a given split, $70\%$ of the scenes constitute the training set, 10\% the validation set and $20\%$ the test set.
Test scores are then averaged over the $5$ repeated experiments.

{\bf Sliding Windows Sampling.}
The samples contained in the datasets $\mathcal{X}^S$, $\mathcal{X}^D$ and $\mathcal{X}^{S+D}$ are constituted by the full history of states and 2D detection associated to the scenes, which span $60$s and present $120$ timestamps each.
Given that our aim is to make predictions based on shorter time intervals, we sample time windows of shorter duration from each scene.
First, we define $5$ values for time window duration: $5$s, $10$s, $15$s, $20$s and $25$s. 
Then, for each window size, we apply a sliding window approach and extract sub-scenes from the whole dataset of the defined time duration. We train the machine learning models on the obtained sub-scenes.
This allows to deeply observe how the scene duration can affect the models performance.

{\bf Vehicle Behavior Classification Results.} 
Fig.~\ref{fig:results_RF-LSTM} shows results achieved by the RF and the LSTM-based model on the behavior classification task. We report as relevant KPIs the following: auROC, area under the precision-recall curve (auPR), accuracy, recall, precision and F1 score.
We train and test the models on both the $\mathcal{X}^S$ and $\mathcal{X}^D$ datasets and show results for all considered time window duration values.

\begin{figure}[t]
    \centering  
    \includegraphics[clip,width =  \linewidth]{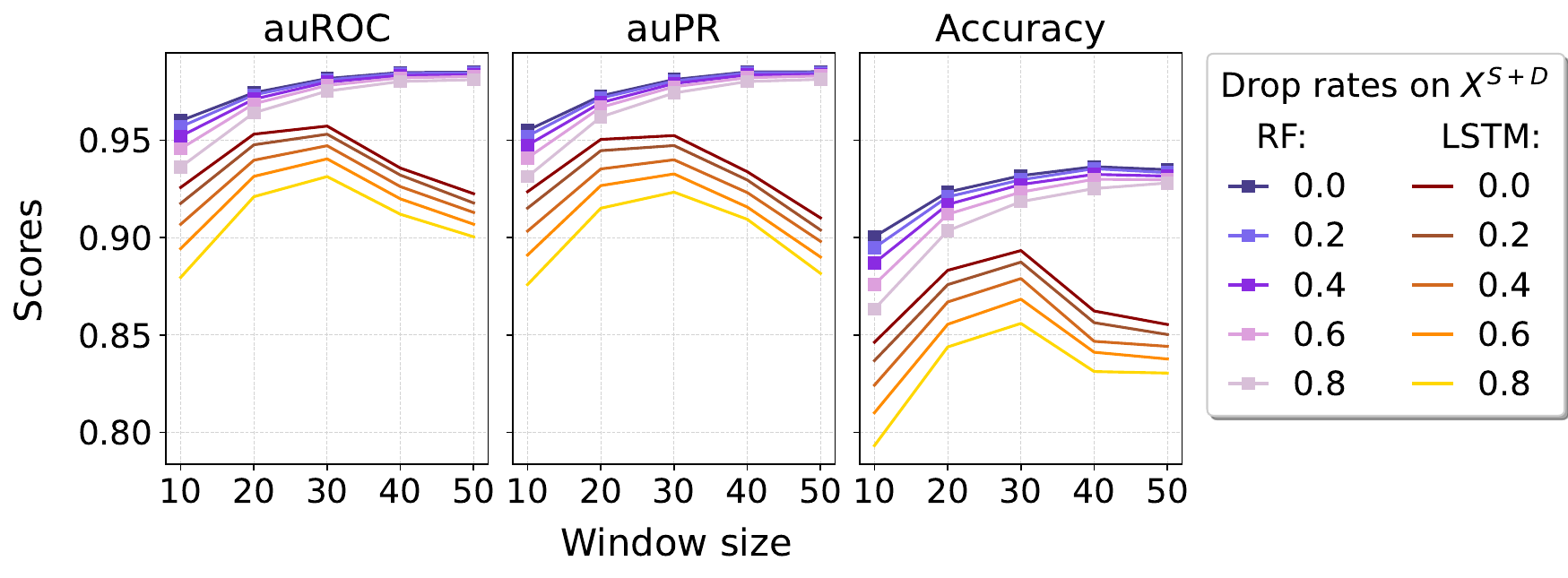}
    \caption{Effect of state information loss across various drop rates. The degradation has been simulated by obfuscating state information from the $X^{S+D}$ dataset with various drop rates. $r=0$ corresponds to the best case scenario where all the data is intact. Test scores are averaged over $5$ repeated experiments.}
    \label{fig:results_dropRates}
\end{figure}

We observe that, as expected, the states allow for better generalization performance than the 2D detection in the pixel coordinates space.
Overall, the RF outperforms the LSTM-based model.
For the RF, we observe that test scores improve as the time window duration increases.
Conversely, the LSTM-based model achieves its peak performance on $30$ timestamps for state-based predictions, and $40$ timestamps for detection-based predictions.
Both models maintain a similar performance gap between $\mathcal{X}^S$ and $\mathcal{X}^D$ across all time window duration values. 

\begin{figure}[t]
    \centering  
    \includegraphics[clip,width =  \linewidth]{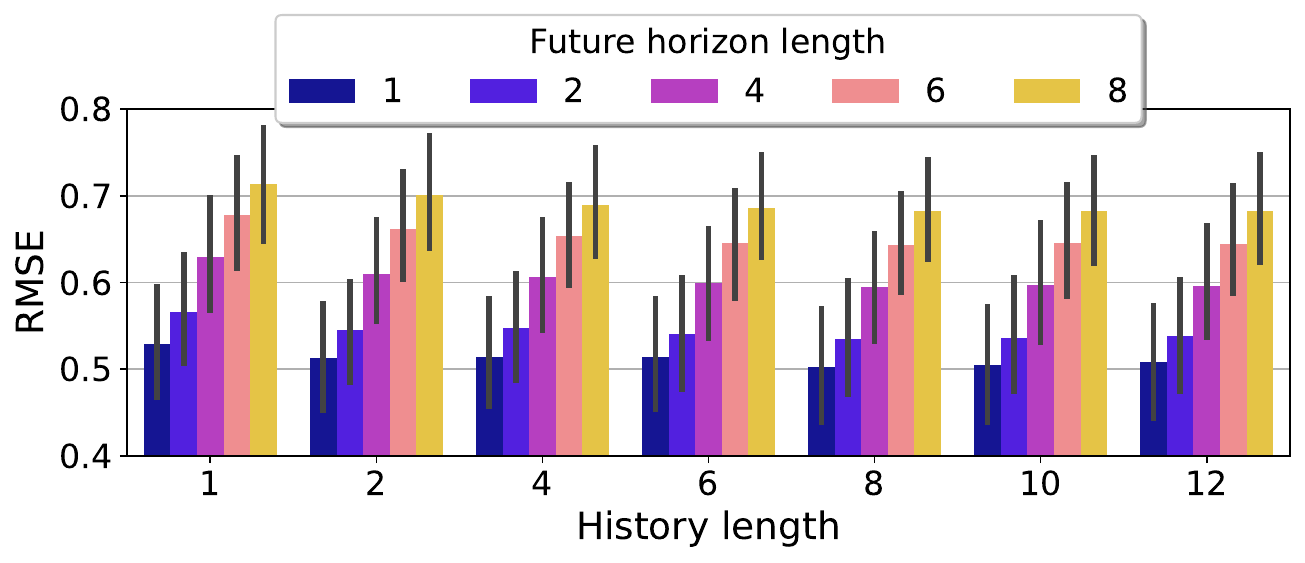}
    \caption{Effect of different history and future horizon lengths. Test results for future state autoregression are averaged over $5$ repeated experiments. Error bars represent standard deviation. RMSE: root mean squared error. Time duration is quantified in terms of timestamp number. $1$ timestamp is approximately $0.5$s.}
    \label{fig:autoregression-1}
\end{figure}

{\bf Effect of Target State Degradation.} 
For this experiment campaign, we degrade the state information from the $\mathcal{X}^{S+D}$  dataset to test the system robustness against sensor failures and/or disruptions in communication, such as delays or disconnections. Please note that 2D detection information is not perturbed.

Given a drop rate $r \in (0,1)$, we randomly obfuscate a portion of rows in $\bm{X}^S$ defined by $r$, while leaving $\bm{X}^D$ unaffected. 
In particular, the state vectors corresponding to the selected rows are set to {\tt NaN} at test time.
Subsequently, for the affected rows, the models consider the latest known state for the classification. 
Fig.~\ref{fig:results_dropRates} highlights the performances of both models in response to different drop rate values $\{0.2, 0.4, 0.6, 0.8\}$. 
The results are compared to the ideal scenario, i.e., $r=0.0$, which represents the upper limit. 
We observe that RF test scores results in a $\sim3\%$ of performance drop. Analogously, the LSTM-based model faces a $\sim5\%$ of performance degradation.

All in all, the car detection task is successfully executed demonstrating the feasibility of our proposed approach while keeping the overall solution complexity affordable. The output of this task can be fed into the autonomous control system pursuing a settings fine-tuning while minimizing the difference between a skilled-human driven behaviors and consistent autonomously driving operations.



%

\subsection{Future State Autoregression}
In the following, we present experimental results for the future state autoregression task. The term ``auto'' in {\it autoregressive} indicates that a variable is regressed against itself~\cite{hyndman2018forecasting}.
We first present a formalization of the problem. Then, we show the effect of different past (i.e., history) and future horizon lengths on the prediction error.
Finally, we compare the performance achieved by the model on scenes wherein the target vehicle (TV) is controlled by the autonomous agent against scenes in which the vehicle is driven by a human. 

{\bf Problem Formulation.} 
Given a history of $H$ past timestamps and a future horizon defined by $F$ timestamps, our goal is to learn an autoregressive parametric function $f_{\bm{\theta}} : \mathbb{R}^{H \times 5} \xrightarrow{} \mathbb{R}^{F \times 5}$ such that
\begin{equation}\label{eq:autoregression}
\hat{\bm{X}}_{H:F}^S = f_{\bm{\theta}}\big(\bm{X}_{1:H}^S\big),
\end{equation}
where $\bm{\theta}$ are learnable parameters and $\bm{X}_{A:B}^S = \big[\bm{x}^S_{i,t_A}, \dots, \bm{x}^S_{i,t_B} \big]^T \in \mathbb{R}^{(B-A) \times 5}$ represents the state vectors of a scene from timestamp A to B.

We train the model by optimizing a mean squared error loss, as follows:
\begin{equation}\label{eq:loss-autoregression}
\min_{\bm{\theta}} L\big(\bm{\theta}\big) = \frac{1}{N} \sum_{i = 1}^{N} \Big(f_{\bm{\theta}}\big(\bm{X}_{i, 1:H}^S\big) - \bm{X}_{i, H:F}^S\Big)^2\:\:.
\end{equation}

{\bf Implementation Details.} 
Specifically, we implement $f_{\bm{\theta}}$ as a multi-layer perceptron (MLP) with 3 layers. We unroll the $\bm{X}_{1:H}^S$ and $\bm{X}_{H:F}^S$ matrices and treat them as 1D vectors. The dimension of the $2$ hidden layers is set to $32$ and $16$ accordingly, while the dimension of the input and output is set to $5H$ and $5F$, respectively. We employ sigmoid activations and train the network with the Adam optimizer with an initial learning rate of $10^{-4}$. We set a maximum of $10,000$ epochs and adopt early stopping with $10$ patience epochs monitoring a validation score.     

{\bf Performance Evaluation and Data Preparation.} 
Analogously to the experiments described above, we perform $5$ independent training/validation/test splits, so that $70\%$ of the scenes is employed at training time, $10\%$ for validation and $20\%$ for test.
After employing the sliding window sampling as described in the previous section, we extract sub-scenes from the full scenes set. For the sampling, we set the window size to $20$ timestamps, i.e. $\sim 10$s.
We train a model for each combination of history length $H \in \{1, 2, 4, 6, 8, 10, 12\}$ and future horizon length $F \in \{1, 2, 4, 6, 8\}$.

\begin{figure}[t]
    \centering  
    \includegraphics[clip,width=\linewidth]{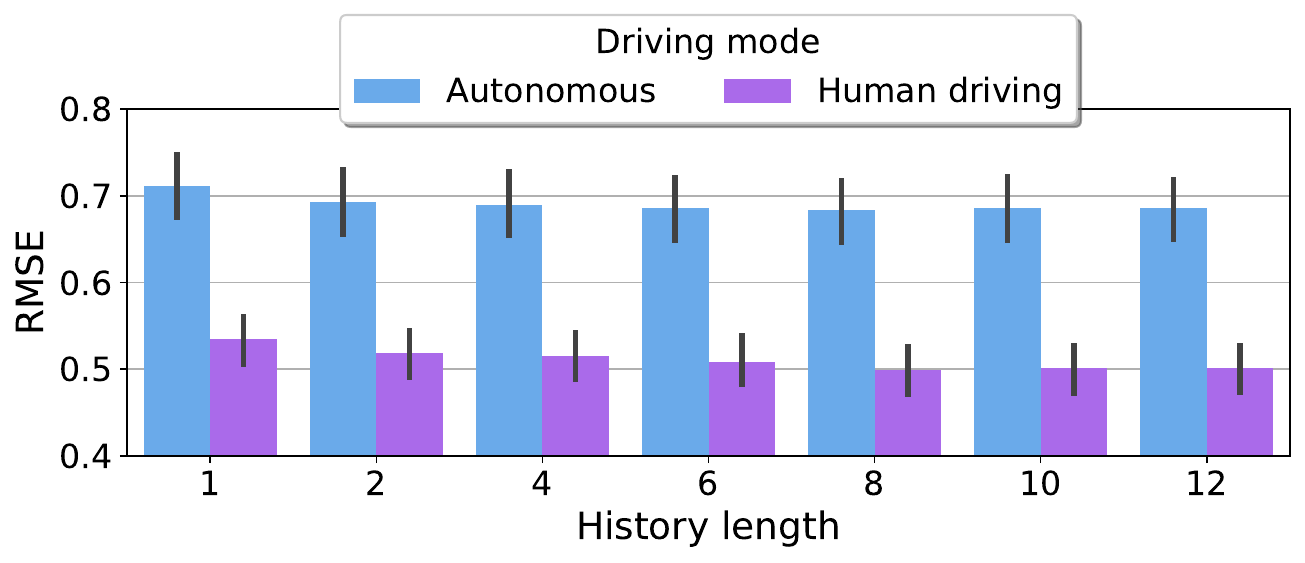}
    \caption{Future state autoregression error for autonomous agent and human driver. Error bars represent the standard deviation over $5$ repeated experiments with different training/test splits.}
    \label{fig:autoregression-2}
\end{figure}

{\bf Effects of History and Future Horizon Lengths.}
We depict in Fig.~\ref{fig:autoregression-1} the corresponding effect of multiple history and future horizon lengths on the prediction error of the TV future state.
We observe that, as the number of timestamps that constitute the future horizon length increases, the prediction error tends to increase. Interestingly, while the input history length increases, the system experiences a little performance improvement, due to more available information to rely on.

{\bf Comparison of Autonomous and Human Driving.}
Finally, in Fig.~\ref{fig:autoregression-2} we compare the prediction error achieved by the same autoregressive model for both autonomous and human driving scenes.
Results are averaged over all considered future horizon lengths.
The results show that the TV future state is significantly easier to predict when a human is driving compared to the case when the Baidu Apollo agent is controlling the vehicle (autonomous).
This is in line with our subjective assessment of the scenes in which the TV is controlled by the autonomous agent. It appears, in fact, that the agent often performs abrupt braking maneuvers and accelerations, which make its overall state harder to predict compared to the smoother driving style of a human driver, as well as less safe due to potential hazards caused by the sudden change of speed and/or direction.

\section{Risk-based Training Process}
An additional result from the comparison described in the previous section is that the lower predictability of the autonomous behavior can result in an increased exposure to the safety risks, as it leads to abrupt braking and/or sudden acceleration or deceleration. This is usually considered among the major causes of hazards for the bystanders and the EV passengers~\cite{IVS2017}, as an exhaustive hazards analysis and risk assessment conducted according to ISO 26262~\cite{iso26262} could highlight. The higher predictability of human driving could result from the inherent human capability to perceive the risk in diverse scenarios~\cite{Kolekar2020}, which suggests the idea to perform an iterative training process of the autonomous driving control algorithm using as objective function the difference between autonomous and human driver prediction errors or similar metrics based on risk estimation, to finally minimize unpredictable behavior effects with subsequent safety risks. This iterative training process is out of the scope of this paper and will be the subject of future works. Future works will also seek to integrate data sourced from real vehicles, validating its suitability for real world applications.

\section{Conclusions}
In this paper, we have presented a novel autonomous vehicle detection solution capable of identifying whether vehicles are driven by humans or computers. We generated and released publicly a custom dataset for this purpose, NexusStreet. It combines both autonomous-driven scenes, obtained on the CARLA simulator by means of Baidu Apollo control agent, and human-driven ones, acquired by using a steering wheel maneuvered by licensed (and skilled) human drivers.
After training various machine learning models, we show that it is possible to automatically detect autonomous vehicles by evaluating video segments. Accuracy and auROC improves to $+90\%$ when combined with histories of the state vector, i.e., position, speed, acceleration and orientation, locally estimated by all installed sensors. Moreover, information as state vector and local classification can be shared among road users through Vehicle-to-everything (V2X)-based infrastructure, further increasing the overall autonomous vehicles detection accuracy to fine-tune the car autonomous control system.

\bibliographystyle{unsrt}
\bibliography{root}

\end{document}